\documentclass{ecai}
\usepackage{graphicx}
\usepackage{latexsym}

\ecaisubmission   

\usepackage{amssymb}
\usepackage{amsmath}
\usepackage{stfloats}
\usepackage[linesnumbered, ruled]{algorithm2e}
\SetKwRepeat{Do}{do}{while}
\SetAlgoLined
\usepackage[switch]{lineno}
\usepackage{color}
\usepackage{mathrsfs}

\begin{document}

\begin{frontmatter}

\title{Hierarchical Task Network Planning for Facilitating Cooperative Multi-Agent Reinforcement Learning}

\author[A]{\fnms{Xuechen}~\snm{Mu}}
\author[B]{\fnms{Hankz}~\snm{Hankui Zhuo}\thanks{Corresponding Author. Email: zhuohank@mail.sysu.edu.cn.}}
\author[C]{\fnms{Chen}~\snm{Chen}} 
\author[A]{\fnms{Kai}~\snm{Zhang}} 
\author[B]{\fnms{Chao}~\snm{Yu}}
\author[C]{\fnms{Jianye}~\snm{Hao}}

\address[A]{Jilin University}
\address[B]{Sun Yat-sen University}
\address[C]{Huawei Noah’s Ark Lab }


\begin{abstract}

Exploring sparse reward multi-agent reinforcement learning (MARL) environments with traps in a collaborative manner is a complex task. Agents typically fail to reach the goal state and fall into traps, which affects the overall performance of the system. To overcome this issue, we present SOMARL, a framework that uses prior knowledge to reduce the exploration space and assist learning. In SOMARL, agents are treated as part of the MARL environment, and symbolic knowledge is embedded using a tree structure to build a knowledge hierarchy. The framework has a two-layer hierarchical structure, comprising a hybrid module with a Hierarchical Task Network (HTN) planning and meta-controller at the higher level, and a MARL-based interactive module at the lower level. The HTN module and meta-controller use Hierarchical Domain Definition Language (HDDL) and the option framework to formalize symbolic knowledge and obtain domain knowledge and a symbolic option set, respectively. Moreover, the HTN module leverages domain knowledge to guide low-level agent exploration by assisting the meta-controller in selecting symbolic options. The meta-controller further computes intrinsic rewards of symbolic options to limit exploration behavior and adjust HTN planning solutions as needed. We evaluate SOMARL on two benchmarks, FindTreasure and MoveBox, and report superior performance over state-of-the-art MARL and subgoal-based baselines for MARL environments significantly.
\end{abstract}

\end{frontmatter}

\section{Introduction}

Deep multi-agent reinforcement learning (MARL) achieves satisfactory results on multiple challenging problems, such as games \cite{mnih2015human}, autonomous driving \cite{lillicrap2015continuous}, etc., which are modeled as multi-agent systems \cite{andriotis2019managing}. Since the development of the simplest method, independent Q-learning (IQL) \cite{Tan1993MultiAgentRL}, which views each agent as an individual, there have been many advanced MARL approaches, such as VDN \cite{sunehag2017value}, QMIX \cite{rashid2018qmix}, CommNet \cite{sukhbaatar2016learning}, DyMA \cite{wang2020few}, COMA \cite{foerster2018counterfactual}, G2ANet \cite{liu2020multi}, MADDPG \cite{lowe2017multi}, MAPPO  \cite{yu2021surprising}, MAVEN \cite{mahajan2019maven}. 
Although those approaches have effectively addressed the difficulties of effective exploration as well as agent communication due to the growth of states and action spaces \cite{rashid2018qmix}, they are all developed based on the assumption of dense rewards.
There are, however, many real-world scenarios with sparse rewards, where the environment does not provide immediate rewards to agents. For example, in a multi-agent soccer game, most of the actions, such as \emph{passing} and \emph{dribbling}, cannot be rewarded, and the environment only gives non-zero rewards when one side scores a goal \cite{jeon2022maser}. Current MARL approaches often fail to to learn policies effectively in this multi-agent setting due to the joint actions of agents affecting the multi-agent system and the lack of non-zero reward drive.

\begin{figure}
\centering
\includegraphics[width=2.3in, keepaspectratio]{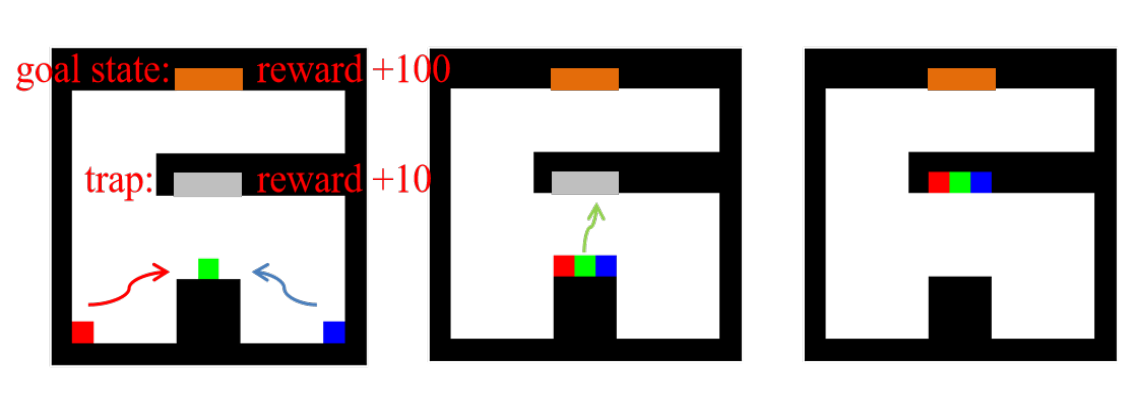}\\
\caption{An illustration of agents falling into a trap and causing the replay buffer to collect no goal state information. The goal of the two agents (red and blue) is to move the box (green block) from the initial state to the target area (orange region).}
\label{fig1}
\end{figure}

To address this issue, one way is to abstract the sparse reward environment and reduce the exploration space of agents, such as the subgoal-based approach, which has been widely used in single-agent environments \cite{kulkarni2016hierarchical,zhang2020generating,yang2018peorl,lyu2019sdrl}. In MARL, a subgoal-based approach for sparse reward was proposed by Tang et al. \cite{tang2018hierarchical}, which is a two-layer hierarchical model (h-MARL) using temporal abstraction to model MARL and incorporating neural networks, where 
higher layer agents learn the assignment of subgoal and lower layer agents explore the environment based on selected subgoals. Yang et al. \cite{yang2019hierarchical} introduced the hidden space into h-MARL such that high-level agents can learn how to select complementary latent skill variables. However, the higher and lower layers of both methods are designed by neural networks, leading to low data efficiency and the lack of interpretability in multi-agent systems. In addition to this class of domain knowledge-based hierarchical methods, Jeon et al. \cite{jeon2022maser} also proposed a replay buffer-based subgoal automatic generation technique (MASER), which avoids the manual design of subgoals. 
However, in some scenarios, agents may trigger traps in the environment and die before exploring the goal state, leading to a lack of information about the target state in the replay buffer. 
As shown in Figure \ref{fig1}, when the box (green block) is moved to the target area (orange area) or the trap (gray area) by two agents (red and blue), the environment will terminate and the two agents will receive corresponding rewards.
As the box is closer to the trap, the agent may give up exploring the target area in order to get a higher reward in a short period of time, which ultimately results in the replay buffer not collecting goal states and not generating valid subgoals.

To address the above issue, we propose a novel architecture that combines MARL with symbolic planning in MAS, called Symbolic Selection for Multi-Agent Reinforcement Learning (SOMARL).
Specifically, we assume there exists a bijective relationship between knowledge and MARL environment, and based on this assumption, we propose a scheme for the construction of symbolic knowledge. Then, we formalize the symbolic knowledge using a hierarchical domain definition language (HDDL) \cite{holler2020hddl} and option framework to obtain the domain knowledge and the symbolic option set, respectively.
Further, the planning module searches for solutions in domain knowledge, and the meta-controller selects appropriate symbolic options from the set of symbolic options to assign to low-level agents according to the planning solution and guides them to explore the environment. Notably, in order for the planning module to take into account low-level environment exploration in its solutions and to constrain the behavior of agents in low-level exploration, we define a new intrinsic reward calculation method in the meta-controller.

We summarize the contributions of this paper as follows:
\begin{itemize}
\item 
To the best of our knowledge, this is the first work to integrate HTN planner and MARL in subgoal-based multi-agent reinforcement learning, 
i.e., applying the HTN to guide MARL and in turn being assisted by MARL in planning.

\item 
We propose a unified knowledge construction scheme to reduce the exploration space of agents and improve the goal success rate of agents in different collaborative sparse reward MARL environments with traps. Moreover, hierarchies with symbolic options have better interpretability than h-MARL.
\item 
We design a novel method of intrinsic reward calculation in the meta-controller. On the one hand, it is used to constrain the exploration behavior of low-level agents. On the other hand, it is used to control the planning solution of the HTN planner.
\end{itemize}

\section{Related Work}
\subsection{MARL with centralized training and decentralized execution}

Centralized training and decentralized execution (CTDE) \cite{oroojlooy2022review,rashid2018qmix} is a key architecture in MARL. The architecture uses the central controller to assist the agents in learning policies during training, each agent makes decisions based on local observations and no longer needs a central controller during execution. VDN \cite{sunehag2017value}, QMIX \cite{rashid2018qmix}, QTRAN \cite{son2019qtran}, etc. are value-based CTDE methods., and the other is policy-based methods, including MADDPG \cite{lowe2017multi}, COMA \cite{foerster2018counterfactual}, MAPPO \cite{yu2021surprising}, etc. MAPPO is an extension of the PPO \cite{schulman2017proximal} on multi-agent scenarios, using an actor-critic architecture, where critic learns a central value function. In this paper, we use MAPPO as the low-level learning algorithm.

\subsection{Subgoal Assignment}

In the single-agent case, Kulkarni et al. \cite{kulkarni2016hierarchical} proposed a Hierarchical Deep Reinforcement Learning (h-DRL) method, which abstractly models the original MDP problem to obtain a subgoal set, and divides the solution process into high-level and low-level. Among them, the high level selects the appropriate subgoal from the subgoal set by meta policy and assigns it to the low-level, and then the low-level learns to explore the environment to reach the subgoal. Although this method successfully solves a class of Atari tasks with sparse and delayed reward signals (e.g.\ Montezuma’s Revenge), it lacks interpretability and data efficiency because of higher and lower level neural network architectures. To alleviate these issues, a class of hierarchical frameworks that integrate knowledge-based symbolic planning with RL was proposed \cite{cimatti2008automated,hogg2010learning,leonetti2016synthesis,yang2018peorl,lyu2019sdrl}. In this paradigm, high-level is required to collect prior knowledge, which is formulated as a formal, logic-based language such as Planning Domain Definition Language (PDDL \cite{McDermott1998PDDLthePD}), and guide low-level agent to explore effectively.

Considering the existence of multi-agent scenarios in the real world, Tang et al. \cite{tang2018hierarchical} extended h-DRL on MARL, in which the hierarchical structure based on Qmix network (h-Qmix) is considered to be the most effective. On this basis, HSD \cite{yang2019hierarchical} introduces skill and extends h-Qmix. In addition, Andrychowicz et al. \cite{andrychowicz2017hindsight} proposed to use replay buffer to define subgoal. MASER \cite{jeon2022maser} extended it to multi-agents and proposed to consider both local Q-value and global Q-value of the agents, making it possible to select a more efficient subgoal from the replay buffer. However, since this method generates subgoals from the replay buffer, in some multi-agent scenarios with traps, the agents may fail to explore the goal state so that the replay buffer cannot generate effective subgoals. On the other hand, the subgoal based on symbolic knowledge makes SOMARL more interpretable on MASs compared to existing methods.

\section{Preliminaries}
In this section, we explain the relevant notation and briefly introduce the key aspects of symbolic knowledge and MARL.

\subsection{Hierarchical Task Network (HTN) Planning with HDDL}

In the paper, we adopt the definition of HTN Planning proposed by Holler et al. \cite{holler2020hddl}. in the form of an extension language to PDDL, called Hierarchical Domain Definition Language (HDDL). Specifically, HDDL is based on a quantifier-free first-order predicate logic $\mathscr{L}$ to describe properties of the world, and its set of instance proposition $P$ is used to represent state $S$. To facilitate the distinction from state in MARL, we call it symbolic state. In contrast to classical planning that focuses on symbolic states, HDDL includes two different types of tasks: primitive tasks (also known as actions) and abstract tasks (also known as compound tasks). It is worth noting that the definition of actions is consistent with PDDL, i.e., $\mathit{a=(name, pre^+, pre^-, eff^+, eff^-)}$, where $\mathit{name}$ is the name of the action model, and $\mathit{(pre^+, pre^-)}$ and $\mathit{(eff^+, eff^-)}$ denote the preconditions and effects of the action model, respectively. 

In particular, an abstract task $t$ is just a task name $X$, i.e., an atom. Its declaration contains two elements: the name X and argument types. For example, (:task deliver :parameters (? r1 - robot \ ?r2 - robot ?p - package)) is a declaration of an abstract task, where "deliver" is the name of the abstract task, and "robot" and "package" are two argument types of the abstract task. Its purpose is not to induce state transition but to reference a pre-defined mapping $M$ to one or more task networks $\mathit{tn}$ by which that abstract task can be refined. A task network $\mathit{tn}$ over a set of task names $X$ is a tuple $(I, \prec, \alpha, VC)$, where $I$ is a set of task identifiers (using $I$ because an abstract task may appear multiple times in the same task network), $\prec$ is a strict partial order over $I$, and $\alpha$ and $VC$ are mappings from $I$ to $X$ (linking task networks to abstract tasks) and a set of variable constraints, respectively. It is worth noting that $I$, $\prec$, and $VC$ in $\mathit{tn}$ can all be empty. A method $m \in M$ is a triple $(X, \mathit{tn}, VC)$, which is used to transform a task network $\mathit{tn}$ to another. Here, $X$ and $VC$ are the name of the abstract task and the set of variable constraints, respectively. It should be noted that the subtasks of a decomposition method $m$ have two keywords: ordered-subtasks (supporting totally ordered) and subtasks (allowing partially ordered tasks), and all abstract tasks and actions in the domain can be used as subtasks.

The HTN planning domain $\mathscr{D} = (\mathscr{L}, T_P, T_C, M)$ contains the predicate logic $\mathscr{L}$, the primitive task set $T_P$, the abstract task set $T_C$, and the set of decomposition methods $M$ used to decompose $T_C$, which implicitly defines all symbolic states $S$. Based on the domain, we further define a planning problem $(\mathscr{D}, s_I, {\mathit{tn}}_I)$, where $s_I$ and ${\mathit{tn}}_I$ represent the initial symbolic state and initial task network, respectively. Finally, solving the HTN planning problem yields a solution $\Pi$ consisting of primitive tasks. Note that if $\Pi$ contains abstract tasks, the subtasks it contains must be executed in order.

\subsection{Multi-Agent Reinforcement Learning (MARL)}

A fully cooperative environment can be described as Decentralized Partially Observable Markov Decision Processes (DEC-POMDP) \cite{oliehoek2016concise}, which is represented by a tuple $\left \langle \widetilde S, \widetilde A, \widetilde O, \widetilde P, \widetilde R, n, \gamma\right \rangle$, where $\widetilde S$ is the state space, $\widetilde A=\widetilde A_1\times \widetilde A_2\times \cdots \times \widetilde A_n $ is the joint action space composed of the action space of all agents, $\widetilde o_i=\widetilde O(\widetilde s;i)$ represents the local observation of the $i$-th agent in the state $\widetilde s$. $P(\widetilde s^{\prime}|\widetilde s, \widetilde a)$ is the state transition probability of $n$ agents from state $\widetilde s$ to ${\widetilde s}^{\prime}$ given $\widetilde a$. $\widetilde R(\widetilde s,\widetilde a)$ represents the shared reward function and $\gamma$ is the discount factor. In a fully cooperative environment, each agent optimizes the common discounted accumulated reward $J(\theta_{1}, \cdots, \theta_{n})=\mathbb{E}_{\widetilde s^t, \widetilde a^t}[\sum_{t}\gamma^t\widetilde R(\widetilde s^t,\widetilde a^t)]$ by its own policy function $\pi_{{\theta}_i}(\widetilde a_i|\widetilde o_i)$.

\subsection{Option Framework}
Hierarchical Reinforcement Learning (HRL) \cite{kulkarni2016hierarchical,nachum2018data,parr1997reinforcement} is a class of methods that models high-level behavior as temporal abstraction actions. We consider a two-layer hierarchical model, and use the option framework to model temporal abstract actions as options \cite{sutton1999between}. Specifically, an option $o=(I_{o}(\widetilde s),\pi_{o}(\widetilde s), \beta_{o}(\widetilde s))$, where $I_{o}(\widetilde s)$ is used as the initial condition of option $o$ to determine whether $o$ is executable at state $\widetilde s$, the termination condition $\beta_{o}(\widetilde s)$ is used to determine whether $o$ is terminable at state $\widetilde s$, and $\pi_{o}(\widetilde s)$ is a policy function that maps state $\widetilde s$ to low-level action. In the option framework, one needs to define the set of options in advance, and agents at higher levels need to learn the policy of choosing the optimal option and assigning it to agents at lower levels, which we call the controller level.

\subsection{Formalization of Multi-Agent Reinforcement Learning (MARL) Environments}


In this section, we define the formal rules for MARL environments based on a tree structure, which serves as the foundation of SOMARL. Specifically, a tree is a tuple $\mathit{TREE}=(V,E)$, where $V$ is the set of nodes and $E \subseteq V \times V$ is a set of edges. It is worth noting that $\mathit{TREE}$ does not contain cycles and has a special node called the root node. For example, consider a tree $\mathit{TREE}=(V,E)$, where $V={a,b,c,d,e,f,g,h,i}$ and $E$ contains the following edges: ${(a,b), (b,c), (c,d), (c,e), (b,f), (f,g), (f,h), (h,i)}$. Here, $a$ is the root node.

Furthermore, we can use HDDL and the option framework to construct planning problems and options on $\mathit{TREE}$, as described in the next section. It is worth noting that since the planning problem and options come from the same $\mathit{TREE}$, it naturally connects HTN planning with the option framework, and $\mathit{TREE}$ acts as a bridge between the two.

\begin{figure}
\centering
\includegraphics[width=2.38in, keepaspectratio]{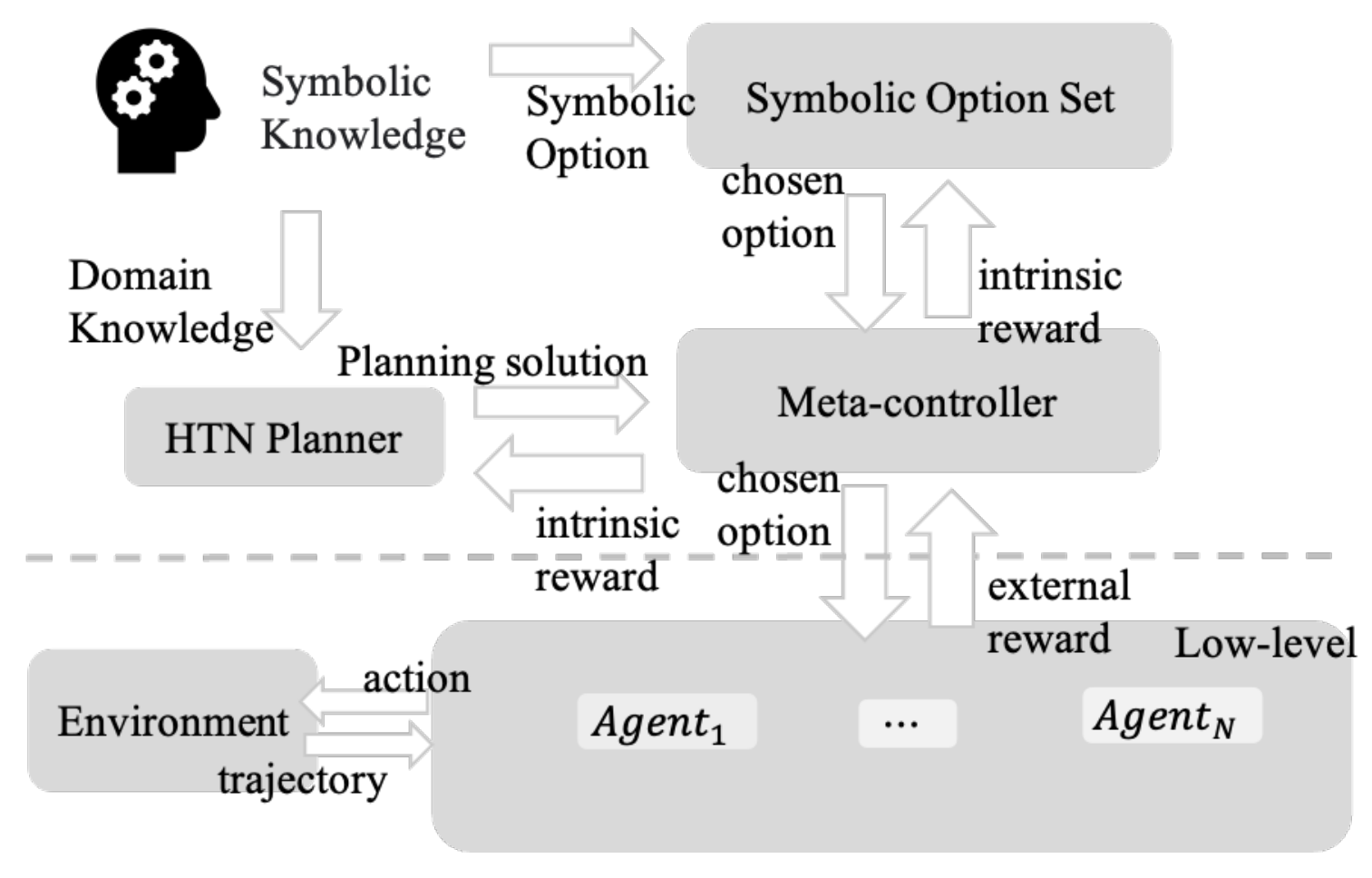}\\
\caption{The SOMARL framework consists of four sub-modules: knowledge, meta-controller, HTN planner, and agents.}
\label{fig2}
\end{figure}

\section{The SOMARL Framework}
We formalize the MARL environment as a tuple\ $(\mathit{TREE}, \mathscr{L}, T_P, T_C, M, s_I, {\mathit{tn}}_I, \widetilde S, \widetilde A,\widetilde O,\widetilde R,\widetilde P, n, \gamma)$, which can be divided into three parts:
\begin{itemize}
\item Firstly, we design the knowledge on the MARL environment in advance (denoted as $\mathit{TREE}$), which is a bijection between HTN planning and option framework.
\item Secondly, $(\mathscr{L}, T_P, T_C, M, s_I, {\mathit{tn}}_I)$ are symbolically described using HDDL on $\mathit{TREE}$.
\item Finally, we define a decision problem using a DEC-POMDP tuple $(\widetilde S, \widetilde A,\widetilde O,\widetilde R,\widetilde P, n, \gamma)$.
\end{itemize}

We consider solving the above problem using SOMARL. Specifically, SOMARL is designed as a closed-loop framework, in which the HTN planner and meta-controller improve the exploration efficiency, and in turn, the low-level policy improves the quality of planning. The SOMARL framework is shown in Figure \ref{fig2}, which consists of four modules: (1) Knowledge: constructs the planning problem and symbolic option set required by HTN planner and meta-controller, respectively, (2) Meta-controller: selects symbolic options based on the plan and calculate intrinsic reward, (3) HTN Planner: generates plans based on domain knowledge and intrinsic rewards of methods, and (4) Multi-agent: interacts with the environment based on the chosen symbolic options. We will address each module in detail below.

\subsection{Construction of Knowledge}

To obtain domain knowledge and the symbolic option set, we first design a formal description $\mathit{TREE}$ for the MARL environment. We define each node $v$ in $\mathit{TREE}$ to correspond to a state in the MARL environment, and each edge describes a change in the state. Specifically, the goal state $s_{goal}$ and the initial state $s_{initial}$ of the MARL environment are defined as the root node and the leaf node of $\mathit{TREE}$, respectively. In addition, we assume that there are enumerable and necessary intermediate states, i.e., subgoals, in the MARL environment from $s_{initial}$ to $s_{goal}$. These subgoals constitute the child nodes of $s_{goal}$, and we can determine the parent-child relationship in $\mathit{TREE}$ based on their temporal order.

\begin{figure}
\centering
\includegraphics[width=3in, keepaspectratio]{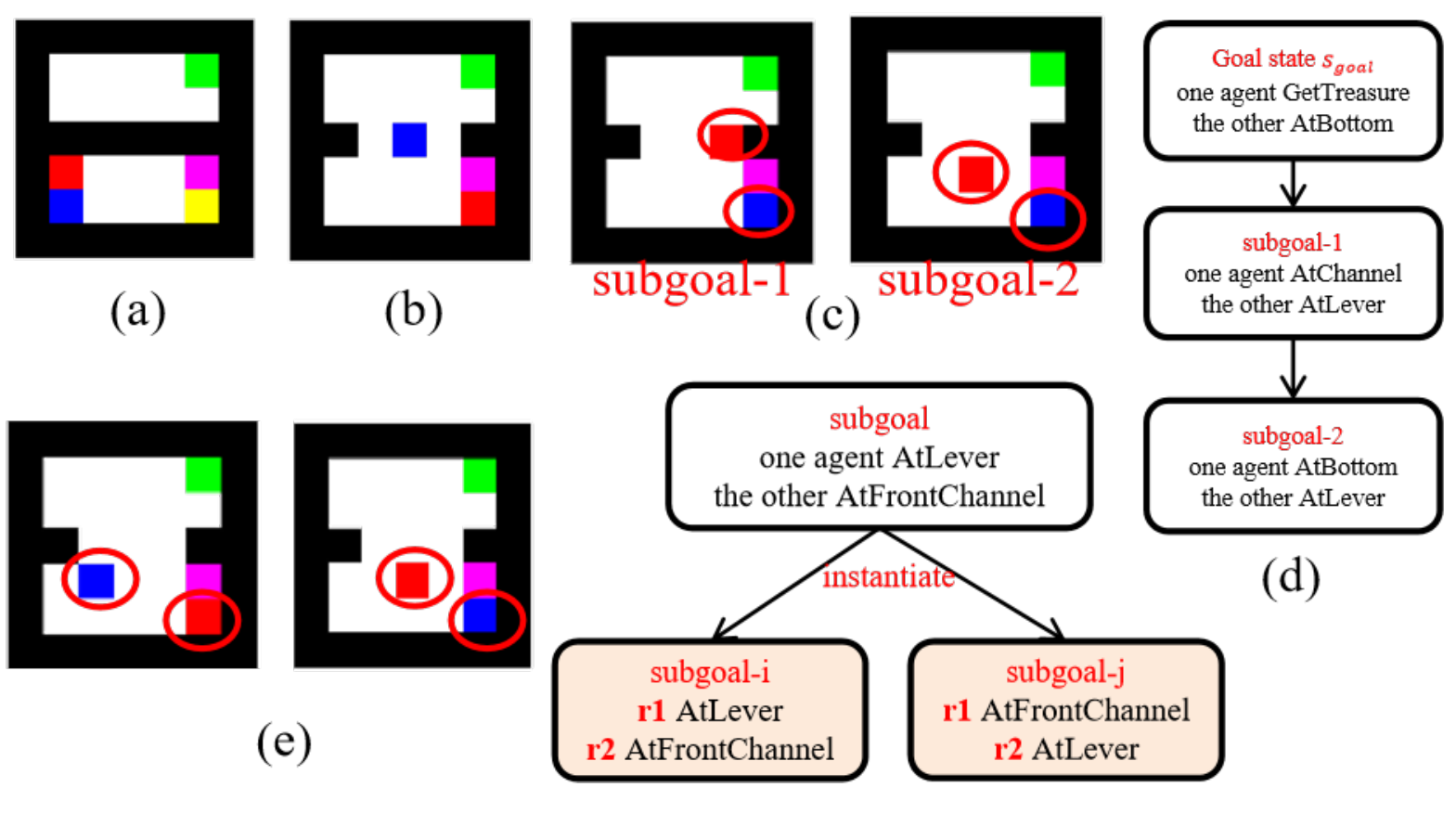}\\
\caption{A figure of building $TREE$ based on the states of a MARL environment. (a) and (b) show the description of the FindTreasure environment, (c) and (d) show instances of vertically expanding $TREE$ based on subgoals, and (e) shows how to instantiate subgoals and horizontally expand $TREE$.}
\label{fig3}
\end{figure}

As shown in Figure \ref{fig3}(a), in the FindTreasure environment, there are two agents (red and blue), a lever (yellow), a trap (pink), and a treasure (green). Only when one agent is at the lever, the other agent can enter the room above (as shown in Figure \ref{fig3}(b)). We set the goal state $s_{goal}$ as when one of the agents reaches the green area, and define two subgoals (as shown in Figure \ref{fig3}(c)): one is that one agent is inside the channel and the other agent is at the lever (subgoal-1), and the other is that both agents are in the room below (subgoal-2). Obviously, subgoal-2 must occur before subgoal-1, so the node representing subgoal-2 is the child of the node representing subgoal-1 (since $s_{goal}$ is the root node and $s_{initial}$ is the leaf node, as shown in Figure \ref{fig3}(d)).

By defining subgoals, we expand $\mathit{TREE}$ vertically. Furthermore, by instantiating each subgoal, we obtain the horizontal expansion of $\mathit{TREE}$, which allows different instances of agents of the same subgoal to correspond to different nodes of the same layer in $\mathit{TREE}$. For example, if the state in which one agent is at the lever and the other agent is at the front of the channel is considered as a subgoal, then such subgoal instantiation corresponds to different tree nodes of the same layer in $\mathit{TREE}$, as shown in Figure \ref{fig3}(e).

In this way,  we can obtain a formal rule $\mathit{TREE}$ that define the path from the initial state to the goal state in the MARL environment.

\subsubsection{Domain Knowledge}

\begin{figure}
\centering
\includegraphics[width=3in, keepaspectratio]{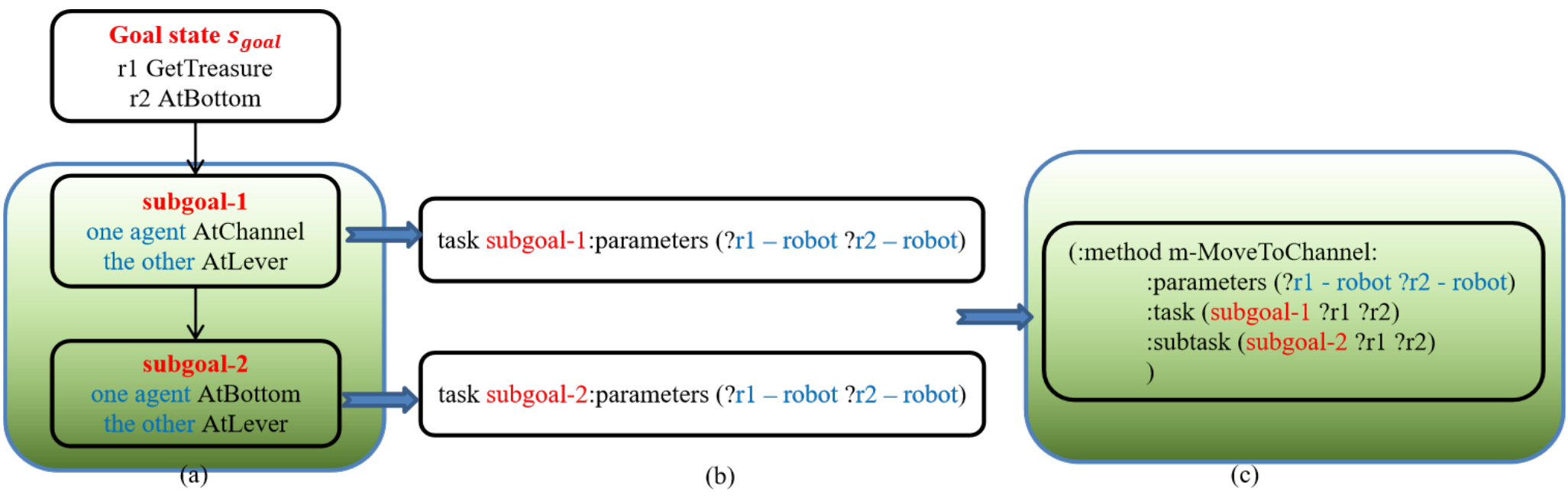}\\
\caption{An example of building domain knowledge on $\mathit{TREE}$. The nodes and edges in $\mathit{TREE}$ correspond to the abstract tasks and decomposition methods in HDDL.}
\label{fig4new}
\end{figure}

After obtaining the symbolic knowledge $\mathit{TREE}$ of the MARL environment, we use the HDDL language to formulate it as $(\mathscr{L}, T_P, T_C, M, s_I, {\mathit{tn}}_I)$. Specifically, we define that the symbolic initial state $s_I$ corresponds to a leaf node in $\mathit{TREE}$, and therefore, a primitive task $T_P$ corresponds to an edge connecting the leaf node. Additionally, in HTN planning, while an abstract task $T_C$ may not involve specific action details (as it describes high-level task), it can be further decomposed into primitive tasks that describe state transitions, so that the ultimate goal of the abstract task $T_C$ is to achieve a certain state. Considering that each child node in $\mathit{TREE}$ corresponds to a subgoal, we can define that the abstract task $T_C$ corresponds to a set of nodes $V$ in $\mathit{TREE}$ (one-to-one correspondence). Furthermore, the instance objects contained in the subgoals of $\mathit{TREE}$ correspond to the argument types of the abstract task $T_C$. As shown in Figure \ref{fig4new}(b), these two abstract tasks correspond to subgoal-1 and subgoal-2 in $\mathit{TREE}$, and their parameter types are the instance objects contained in subgoal-1 and subgoal-2.

In addition, to establish a one-to-one correspondence between the edges $e \in E$ in $\mathit{TREE}$ and the decomposition methods $m \in M$ in HTN planning, we define that the task network $\mathit{tn}$ of each decomposition method $m$ is a single abstract task, and the variable constraint $VC$ and order $\prec$ in $m$ are empty. This definition enables the subtask in $m$ to be determined by the parent-child relationship in $\mathit{TREE}$. As shown in Figure \ref{fig4new}(c), the abstract task corresponding to subgoal-2 is a subtask of the abstract task corresponding to subgoal-1 (since subgoal-2 is a child node of subgoal-1). In particular, the goal state (the root node of $\mathit{TREE}$) corresponds to the initial abstract task ${\mathit{tn}}_I$. 

This approach of constructing the corresponding HTN planning domain from $\mathit{TREE}$ ensures that each path from the leaf nodes (the symbolic initial state $s_I$) to the root node (the initial abstract task ${\mathit{tn}}_I$) of $\mathit{TREE}$ is one of the solutions in HTN planning, which includes both primary tasks and abstract tasks.

\subsubsection{Symbolic Option Set}


In this paper, we extend the symbolic option \cite{jin2022creativity} to multi-agent scenarios and use it to associate HTN planning with MARL. A symbolic option is defined as $(s, \pi, s^{\prime})$, where $s$ and $s^{\prime}$ represent different subgoals, which correspond to two nodes of the same edge in $\mathit{TREE}$ (e.g., subgoal-1 and subgoal-2 in Figure \ref{fig3}c). $\pi$ is a low-level policy that interacts with the environment and is trained with intrinsic reward. Considering that a symbolic option represents the transition of subgoals and HDDL decomposition methods $M$ can correspond to each edge $e \in E$ in $\mathit{TREE}$, we use $\mathit{TREE}$ as a bridge to establish a one-to-one correspondence between symbolic options and decomposition methods $M$. Obviously, the number of edges in $\mathit{TREE}$ is equal to the number of decomposition methods $m$ in HDDL, which is also equal to the number of symbolic options. We use equations \eqref{eq1} and \eqref{eq2} to determine whether the initial condition $I_{so}(\widetilde s)$ and the termination condition $\beta_{so}(\widetilde s)$ hold at state $\widetilde s$.

\begin{equation}
\label{eq1}
I_{so}(\widetilde s)=\left\{
\begin{aligned}
True & , & s=F(\widetilde s), \\
False & , & otherwise.
\end{aligned}
\right.
\end{equation}

\begin{equation}
\label{eq2}
\beta_{so}(\widetilde s)=\left\{
\begin{aligned}
True & , & s^{\prime}=F(\widetilde s), \\
False & , & otherwise.
\end{aligned}
\right.
\end{equation}

It is worth noting that the symbolic option makes SOMARL more interpretable than the black-box neural network-based algorithms.

\subsection{Meta-controller}
In SOMARL, the meta-controller is responsible for selecting the symbolic option to assign to the agents according to the planning solution, and calculating the intrinsic reward of the symbolic option after the agents have finished interacting with the environment.

\subsubsection{Symbolic Option and Intrinsic Reward}

According to the construction of knowledge, the elements of the method set $M$ in HTN planning have a one-to-one correspondence with the elements in the symbolic option set $O$ (since they both correspond to edges $e \in E$ in $\mathit{TREE}$). 
Furthermore, we can use the mapping between the method set $M$ and the symbolic option set $O$ to select symbolic options from the planning solution. Specifically, the planning solution consists of both primitive tasks and abstract tasks, and we first use the mapping $alpha$ in the task networks ${\mathit{tn}}_I$ (all of which are single abstract tasks) to correspond these abstract tasks to decomposition tasks. Then, we construct a new sequence of tasks by using the parent-child relationships in $\mathit{TREE}$ as the order between decomposition tasks and primitive tasks. Finally, the meta-controller selects the corresponding option from the symbolic option set in the order of the new sequence of decomposition tasks and primitive tasks.

In order to better train the policy of symbolic option, we design an intrinsic reward to constrain the transition of $s$ to $s^{\prime}$ in symbolic option, see equation \eqref{eq3}:

\begin{equation}
\label{eq3}
r_i=\left\{
\begin{aligned}
r_e + \phi - n \cdot c & , & \beta_{so}(\widetilde s)=True, \\
0 & , & otherwise.
\end{aligned}
\right.
\end{equation}

Where, $r_i$ is a non-zero value if and only if the termination condition of $so$ is true. $\phi$ and $c$ are hyperparameters of SOMARL, which generally need to be tuned on different MARL environments. Specifically, $\phi$ represents the extra reward received by low-level agents when 
$\beta_{so}(\widetilde s)=True$, so it is a positive number. $c$ is the penalty term coefficient. In the paper, we take $\phi=5$, $c=0.01$ on two environments. 
$r_e$ and $n$ represent the external reward and the reduced count of external reward after the agents perform an action respectively. We specify that when $r_e<0$, $n=1$, otherwise $n=0$.

In addition, in order to improve the interaction efficiency of low-level agents, we use the intrinsic reward of the symbolic option to correct the planning solution of the HTN planner.

\subsection{HTN Planner}


Considering that the purpose of an HTN planner is to assist low-level agents in exploration, the planning solution obtained by the planner should not only have a small cost function value but also take into account the completion of symbolic options by the low-level agents during actual execution. The completion of symbolic options can be evaluated using the intrinsic reward of symbolic options. Therefore, we improve the existing HTN planner to consider the intrinsic rewards of low-level agents when searching for solutions.

\subsubsection{Improved Heuristic Function}
pyHIPOP is an HTN planner that uses heuristic functions to drive the search for solutions in plan space \cite{lesire2020pyhipop}. 
Specifically, in pyHIPOP, the additive heuristic function $h_{madd}(n)$ for a method $m$ calculates the cost of the method. It is composed of the maximum additive heuristic function $h_{add}^{max}(n)$ \cite{bercher2017admissible} of the primitive tasks in the method and the heuristic function $h_{f}(n)$ associated with the method, i.e., $h_{madd}(n)=h_f(n)+max_{n^{\prime} \in Sub(n)}h_{add}^{max}(n^{\prime})$. Here, $Sub(n)$ denotes the set of child nodes of $n$, and $h_{f}(n)$ can be any heuristic function associated with the method, such as the number of primitive tasks in $O_{i}$ or the length of the critical path. We extend $h_{madd}(n)$ as follows:

\begin{equation}
\label{eq4}
\begin{aligned}
h_{madd}(n)=h_f(n)+max_{\prime{n} \in Sub(n)}h_{add}^{max}(\prime{n}) - R_m^k
\end{aligned}
\end{equation}

Where $R_m^k=\sum_{j=1}^{k}r_i$ indicates the cumulative intrinsic reward of the symbolic option corresponding to method $m$ up to the $k$-th episode. At the initial moment, the intrinsic reward of each symbolic option are initialized to $0$.

The improvement enables $h_{madd}(n)$ in pyHIPOP to not only consider the cost of primitive tasks in the method and the contextual information related to the method but also take into account the success of the symbolic option's policy $\pi$ in guiding the low-level agent to reach $s^{\prime}$. The better the policy $\pi$ of the symbolic option, the higher its intrinsic reward, and the smaller $h_{add}^{max}(n)$ will be. We name the HTN planner that solves planning solutions using this novel heuristic function pyHIPOP+.

\subsection{Planning and Learning}

We need to pre-design the knowledge of MARL environment. Specifically, we define the subgoal set on the MARL environment, and instantiate their agents to obtain a symbolic knowledge $TREE$ with the goal state of environment as the root node. Then we use HDDL to formulate $TREE$ to get domain knowledge $D$ and use the correspondence between symbolic option and the edges in $TREE$ to construct symbolic option set $O$. Finally, we define the cumulative intrinsic reward for each symbolic option and their dictionary is denoted as $R$. 
Note that the initial value of each symbolic option in $R$ is $0$.

As shown in Algorithm $1$, SOMARL takes domain knowledge $D$, symbolic option set $O$ and cumulative intrinsic reward dictionary $R$ of symbolic options as input. When an episode $k$ starts, we first get the initial state ${\widetilde s}_0$ of the environment. After that pyHIPOP+ receives the updated cumulative intrinsic reward dictionary $R$ from the previous episode to solve the domain knowledge $D$ to get planning solution $\Pi_k$. Next, the meta-controller selects elements from symbolic option set $O$ to assign to low-level in order of method in $\Pi_k$, combining the correspondence between symbolic option and method, until the environment terminates or the selected symbolic option is reached. 
Note that under each selected symbolic option ${so}_m$, low-level multi-agents will select actions according to the policy of ${so}_m$, and get external reward $r_e$ and the reduced count of external reward $n$ after interacting with the environment. Then meta-controller calculates the intrinsic reward $r_i$ according to the formula \eqref{eq3}, and trains the policy of ${so}_m$ through $r_i$ and interaction data. Finally, the cumulative intrinsic reward of ${so}_m$ is updated by $r_i$ and used for the planning solution at the next episode.

\begin{algorithm}[!ht]
  \label{SOMARL}
  \caption{SOMARL}
  \SetKwInOut{Input}{Input}\SetKwInOut{Output}{Output}
  \Input {domain knowledge $D$, symbolic option set $O$, intrinsic reward  dictionary $R=\{so_0: 0, so_0: 0, ..., so_N: 0\}$}
  \For{$t=1, 2, ..., \mathit{Max\_episodes}$}{
    env.reset()\\
    $\Pi_k \leftarrow \mathit{pyHIPOP\text{+}.solve}(D, R)$\\
    \While{(env isn't terminal) and (steps $<$ Max\_Steps)}{
        $\mathit{{so}_m} = \mathit{meta\text{-}controller}(\Pi_k, O)$ \\
        
        \While{(env isn't terminal) and (steps $<$ Max\_Steps) and ($\beta_{{so}_m}(\widetilde s)=False$)}{
        $\widetilde o, r_e, \widetilde o_{next}, n \leftarrow \mathit{Multi\text{-}agents}({so}_m)$ \\
        $r_i = \mathit{IntrinsicReward}(r_e, n)$ \\
        Training ${so}_m$ with $\widetilde o, \widetilde o_{next}$ and $r_i$ \\
        $R[{so}_m]^k = R[{so}_{m}]^{k-1} + r_i$
        }
    }
   }
\end{algorithm}


\section{Experiments}

In this section, we evaluate SOMARL on two collaborative multi-agent sparse reward environments with traps, FindTreasure and MoveBox. In addition to comparing with baseline multi-agent algorithms (VDN \cite{sunehag2017value}, QMIX \cite{rashid2018qmix}, DyMA \cite{wang2020few}, MAVEN \cite{mahajan2019maven}, CommNet \cite{sukhbaatar2016learning}, COMA \cite{foerster2018counterfactual}, G2ANet \cite{liu2020multi}, MADDPG \cite{lowe2017multi}, MAPPO \cite{yu2021surprising}), we also compared with two subgoal-based algorithms designed for multi-agent sparse reward. One of the algorithms is MASER that defines the subgoal by the replay buffer \cite{jeon2022maser}, and the other is the hierarchical multi-agent reinforcement learning algorithm (h-MAPPO) designed on the same subgoal set and low-level training procedure as our SOMARL.

\begin{figure}
\centering
\includegraphics[width=3in, keepaspectratio]{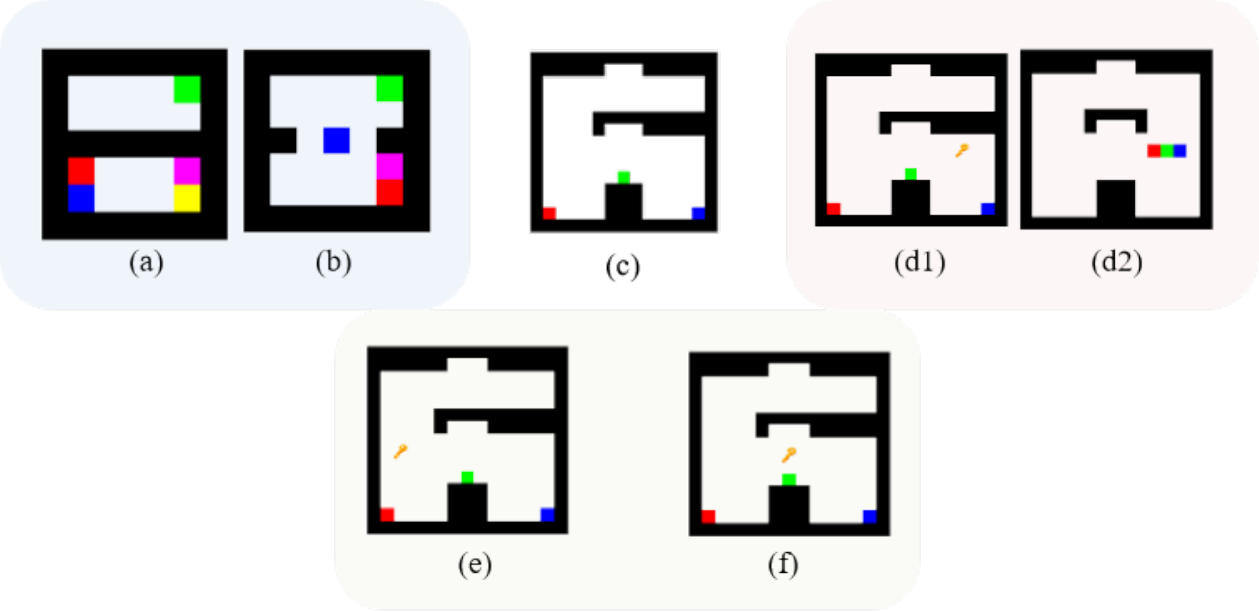}\\
\caption{Task design on two environments, FindTreasure and MoveBox. (a) is a task on the FindTreasure environment, while (c), (d1), (e), and (f) are four tasks on the MoveBox environment.}
\label{fig5new}
\end{figure}

\subsection{Experiments in the FindTreasure Environment}

FindTreasure \cite{jiang2021multi} contains two rooms. At the initial moment both agents (red and blue) are in the lower room and one treasure (green) is in the upper room, their positions are shown in Figure \ref{fig5new}(a). In this environment, if and only if one of the agents reaches the lever (yellow), the other agent can reach the upper room through the channel (as shown in Figure \ref{fig5new}(b)). If an agent reaches treasure, then each agent's reward is $+100$, and the environment terminates. It is worth noting that there is a trap (pink) in FindTreasure, i.e. if one agent is at the lever and the other agent is at the trap, the environment will also be terminated and each agent's reward will be increased by $3$. In addition, if agents hit an obstacle (black) once during the movement, their reward is $-0.1$. The action space of FindTreasure contains five elements: up/down/left/right/wait, and the observation space consists of the position coordinates of two agents.

\begin{figure}
\centering
\includegraphics[width=3.5in, keepaspectratio]{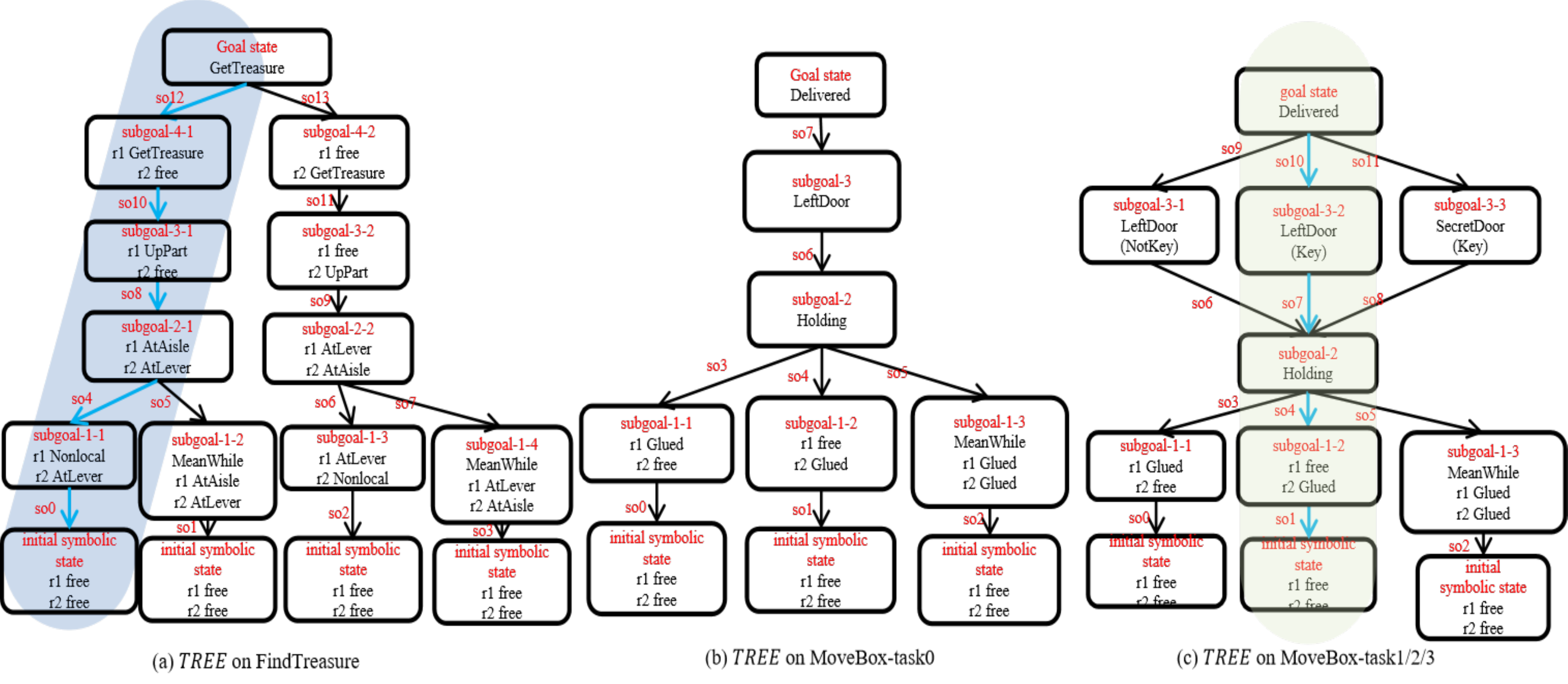}\\
\caption{Design examples of the symbolic knowledge $\mathit{TREE}$ for tasks in FindTreasure and MoveBox environments, where the nodes of $\mathit{TREE}$ represent subgoals and edges represent symbolic options. Specifically, the three tasks in the MoveBox environment (task1/2/3) share the same $\mathit{TREE}$.}
\label{fig6new}
\end{figure}

\subsubsection{Results}

\begin{figure}
\centering
\includegraphics[width=3.3in, keepaspectratio]{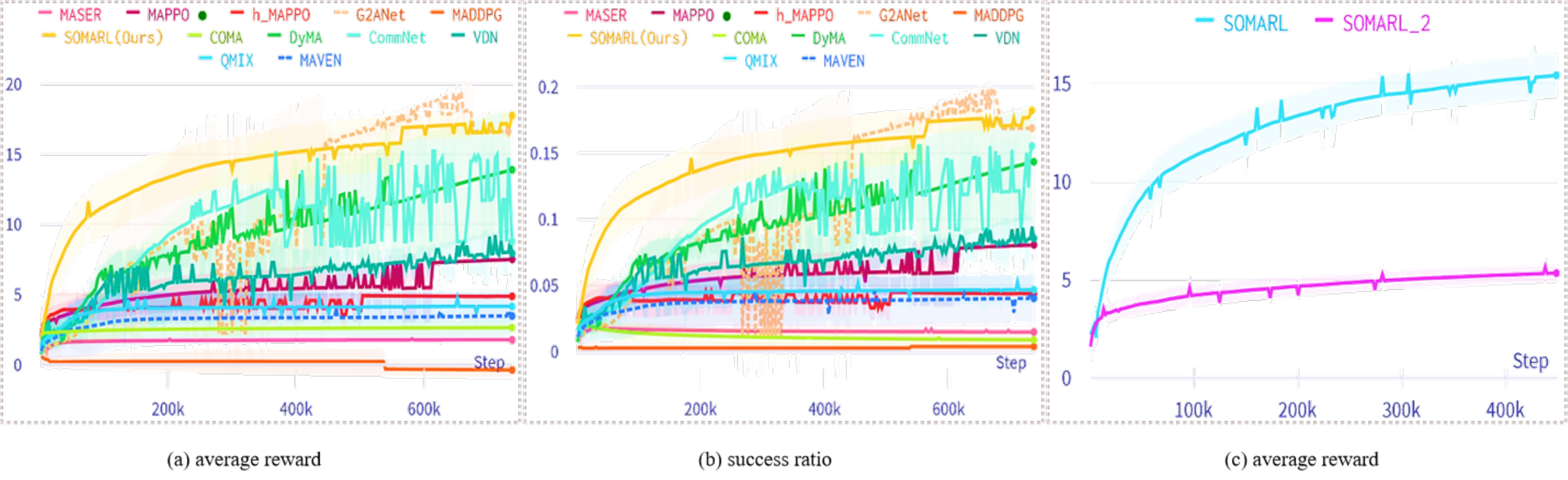}\\
\caption{SOMARL and the performance of various baselines on the FindTreasure environment are shown in the following figures. (a) and (b) show the cumulative average reward and success rate in reaching the goal state during the interaction process of each algorithm, respectively. (c) shows the ablation experiment of intrinsic reward in SOMARL, where the y-axis represents the cumulative average reward value.}
\label{fig5res}
\end{figure}

As shown in Figure \ref{fig6new}(a), we design the symbolic knowledge of FindTreasure and evaluate SOMARL on this basis.
\begin{itemize}
\item  \textbf{Data-efficiency} As shown in Figure \ref{fig5res}(a), SOMARL can get higher rewards faster at the same step. It is noted that although G2ANet achieves higher rewards over time, this method is the least stable. In contrast, SOMARL has the smoothest learning curve among all methods and achieves the highest rewards rewards at the final moments.
\item \textbf{Interpretability} As shown in Figure \ref{fig6new}(a), the $TREE$ consisting of subgoals and symbolic options describes the decomposition of the target task, and its combination with the planning solution can enhance the interpretability of SOMARL. For example, the symbolic option sequence corresponding to the planning solution at the 29th episode is $[{so}_0, {so}_4, {so}_8, {so}_{10}, {so}_{12}]$ (the blue path in Figure \ref{fig6new}(a)). When $r_1$ is not at the trap and $r_2$ is at the lever, the termination condition of ${so}_0$ is successfully reached by low-level, then the meta-controller assigns ${so}_4$ to low-level until $r_1$ is at the channel and $r_2$ is at the lever, which triggers the termination condition of ${so}_4$.
\item \textbf{Success Rate} Figure \ref{fig5res}(b) shows the success rates of different methods to reach the goal state. From the figure, it can be seen that SOMARL eventually achieves a higher success rate and the exploration is overall smooth and less volatile. This suggests that the symbolic knowledge-based approach helps MARL reduce the exploration space of agents.
\end{itemize}


\subsubsection{Ablation Study}

We conduct ablation study on intrinsic reward to verify its effect in SOMARL. From Figure \ref{fig5res}(c), we can see that the exploration ability of SOMARL\_2 decreases significantly after removing the intrinsic reward of symbolic option compared to the original version. This indicates that the intrinsic reward of the symbolic option enhances the learning ability of the model, and the penalty term in the intrinsic reward constrains behavior of low-level agents.


\subsection{Experiments in the MoveBox Environment}

The second environment we use is called MoveBox \cite{jiang2021multi}. As shown in Figure \ref{fig5new}(c), there is a black base in the MoveBox with a box (green) on it, and two agents (red and blue) are located on both sides of the base. The goal of both agents is to move the box to the top area, at which point each agent's reward is $+100$ and the environment terminates. Note that there is a trap in the middle area of MoveBox, when two agents move the box to that area, their respective rewards are $+10$, and the environment is terminated.

In addition, in order to increase the difficulty of reaching the target area for the agents and the box, we add key entities at three locations near the trap (noted as task1, task2, and task3, as in Figures \ref{fig5new}(d1), (e), (f)). When two agents carry the box to get the key, each agent's reward $+5$ and an additional channel to the target area appears (as in Figure \ref{fig5new}(d2)).

\begin{figure}
\centering
\includegraphics[width=3in, keepaspectratio]{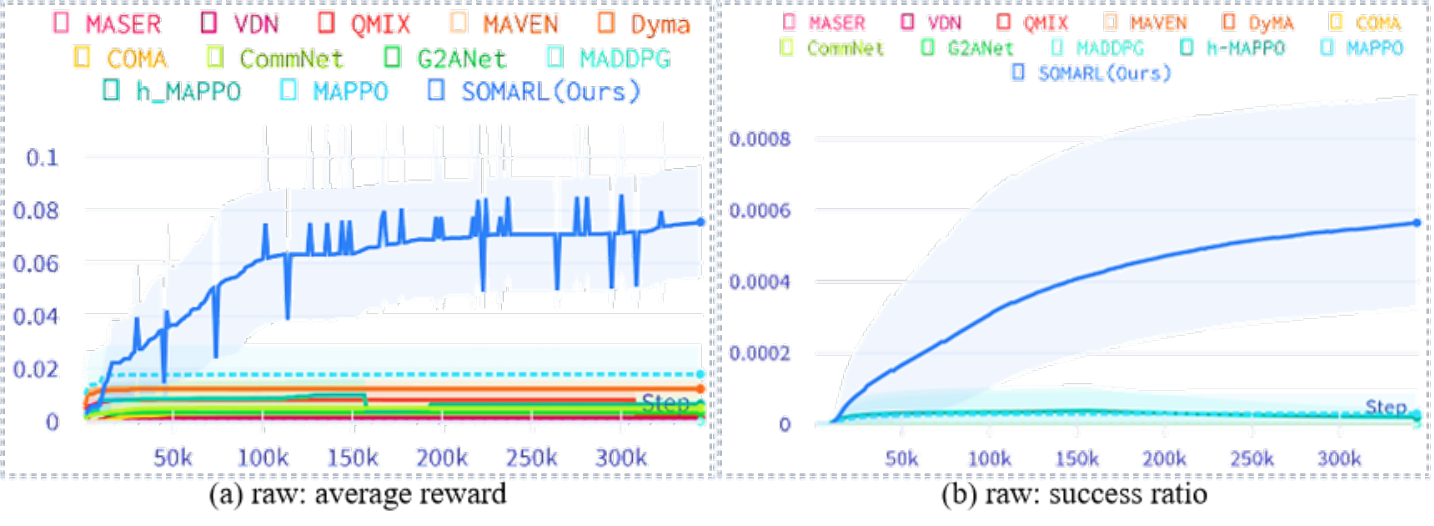}\\
\caption{SOMARL and various baselines' results on task-0 in the MoveBox are shown in this figure. Panels (a) and (b) display the cumulative average reward and success rate in reaching the goal state during the interaction process, respectively.}
\label{figresMoveBoxTask0}
\end{figure}

\subsubsection{Results}

\begin{figure}
\centering
\includegraphics[width=3in, keepaspectratio]{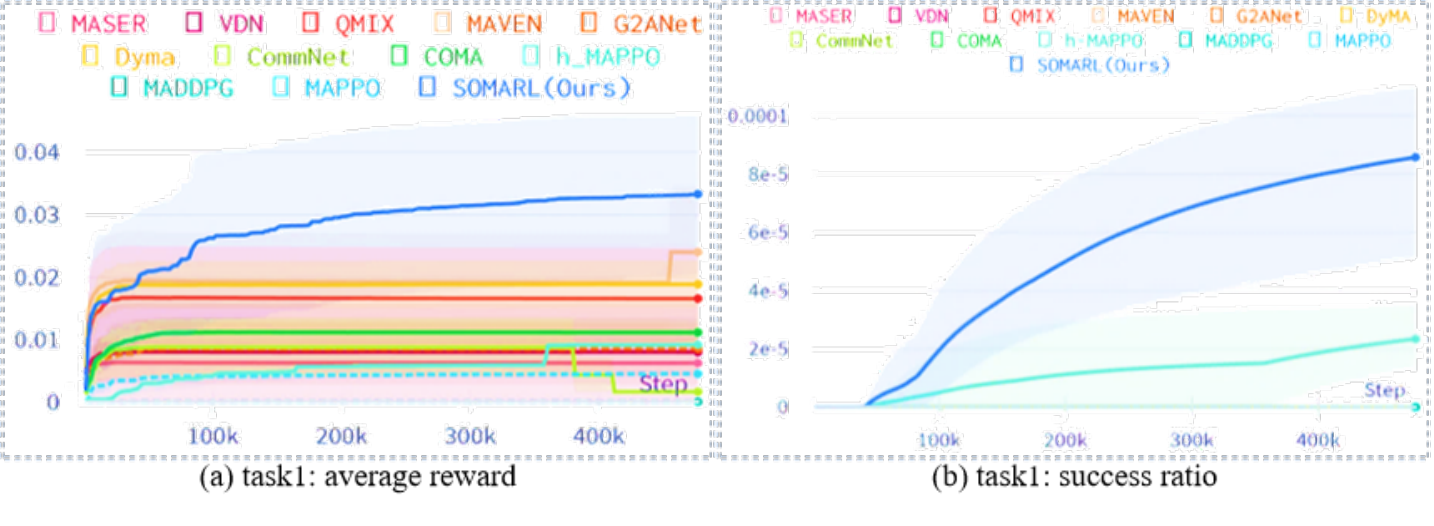}\\
\caption{SOMARL and various baselines' experimental results on task-1 in the MoveBox. The cumulative average reward obtained during the interaction process and the success rate of reaching the target state are shown in (a) and (b), respectively, for each algorithm.}
\label{figresMoveBoxTask1}
\end{figure}

\begin{figure}
\centering
\includegraphics[width=3in, keepaspectratio]{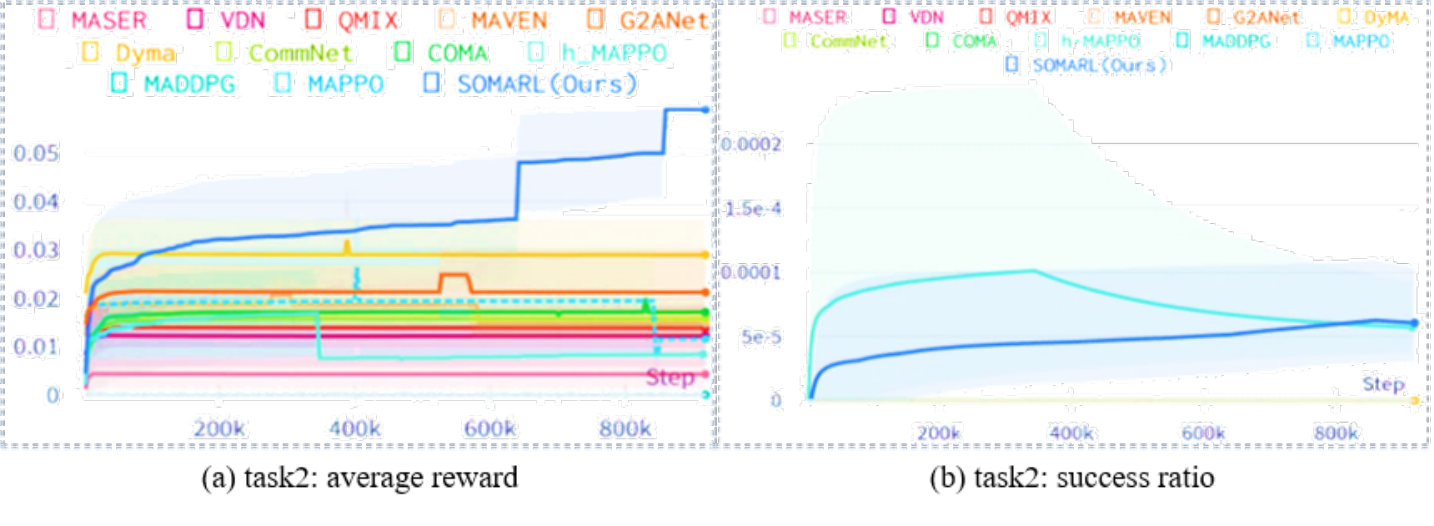}\\
\caption{SOMARL and baselines' results on task-2 in MoveBox environment are shown below. (a) and (b) are the accumulated average reward and success rate of reaching the goal state during the interaction process for each algorithm.}
\label{figresMoveBoxTask2}
\end{figure}

\begin{figure}
\centering
\includegraphics[width=3in, keepaspectratio]{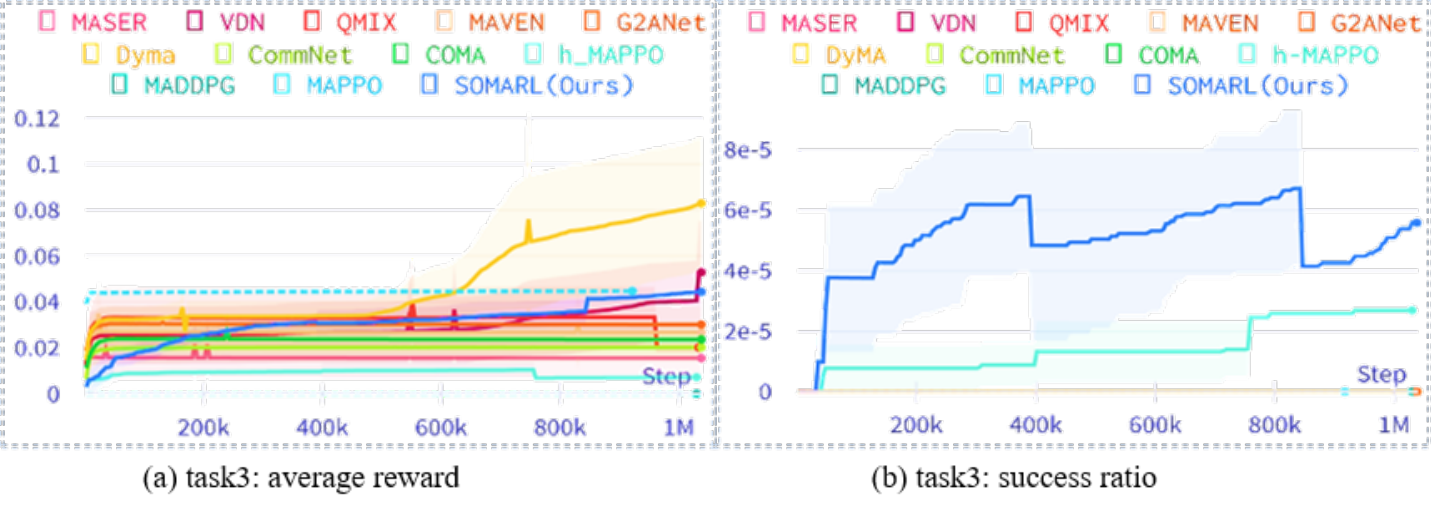}\\
\caption{SOMARL and baselines' experimental results on task-3 in MoveBox environment are shown in (a) and (b), respectively.}
\label{figresMoveBoxTask3}
\end{figure}

As in Figures \ref{fig6new}(b), (c), we design the symbolic knowledge of MoveBox. Notice that although the keys have different placements, they all have the same symbolic knowledge. This shows that symbolic knowledge is only related to the number of subgoals, not to the trajectory of low-level agents, and thus has some generalization. The performance of different methods on different versions of MoveBox is shown in Figure \ref{figresMoveBoxTask0}-\ref{figresMoveBoxTask3}. 

On the whole, SOMARL achieves higher rewards faster in each scenario than other methods, far ahead of those baselines (see Figures \ref{figresMoveBoxTask0}(a), \ref{figresMoveBoxTask1}(a), \ref{figresMoveBoxTask2}(a), \ref{figresMoveBoxTask3}(a)). Combining their success rate, only SOMARL is able to reach the goal state consistently (see Figures \ref{figresMoveBoxTask0}(b), \ref{figresMoveBoxTask1}(b), \ref{figresMoveBoxTask2}(b), \ref{figresMoveBoxTask3}(b)). This is because MoveBox has a stronger trap design compared to FindTreasure, which causes all baselines to terminate exploration early. 

Further, because the key is placed in front of the trap in MoveBox-task3(see Figure \ref{fig5new}(f)), this cause all baselines to forgo exploration and simply choose to take the key and enter the trap to end the interaction. Although this short-sighted behavior makes some baselines gain higher rewards in the short term, it is clear that such exploration by agents is not acceptable. In addition, symbolic knowledge makes SOMARL better interpretable in different scenarios. For example, the symbolic option sequence corresponding to the planning solution in the 6th episode is $[so_1, so_4, so_7, {so}_{10}]$ (the blue path in Figure \ref{fig6new}(c)). When r1 and r2 take the box together, $so_4$ terminates. Then the meta-controller continues to assign $so_7$ to low-level, instructing agents to explore the environment until agents get the key and enter the left channel.

\section{Conclusion}
In summary, we propose a novel multi-agent framework that combines HTN planning and MARL for cooperative multi-agent sparse reward environments with traps. The framework leverages domain knowledge and symbolic options to construct a method for generating symbolic knowledge on the MARL environment. The HTN planner solves the domain knowledge, and the meta-controller selects the symbolic option from the symbolic option set to guide the exploration of the agents. During exploration, the meta-controller computes intrinsic reward to constrain agent behavior and guide the HTN planner. Our experiments demonstrated that SOMARL outperforms existing methods in terms of performance, interpretability, and success rate stability. 

To further enhance the universality of SOMARL, it would be interesting to investigate automatically learning (HTN) domain knowledge \cite{DBLP:conf/aips/ZhuoYPL11,DBLP:journals/ai/ZhuoM014} and planning with the learnt knoweldge \cite{DBLP:journals/ai/ZhuoK17} for facilitating multi-agent reinforcement learning. It would also be interesting to investigate the integration between plan recognition \cite{DBLP:journals/tist/Zhuo19,DBLP:journals/tist/ZhuoZKT20} and multi-agent reinforcement learning.

\bibliography{ecai}
\end{document}